\pdfoutput=1

\documentclass[11pt]{article}

\usepackage[]{acl}

\usepackage{times}
\usepackage{amsmath}
\usepackage{amsfonts}
\usepackage{latexsym}

\usepackage[T1]{fontenc}

\usepackage[utf8]{inputenc}

\usepackage{microtype}

\usepackage{graphicx}
\usepackage{booktabs}
\usepackage{floatrow}
\usepackage{booktabs}
\usepackage{numprint}
\usepackage{subcaption} 
\usepackage{enumitem}
\usepackage{todonotes}
\usepackage{multirow}
\usepackage{arydshln}
\usepackage{bm}

\newcommand{\ahmet}[1]{\textcolor{black}{ #1}}
\newcommand{\peft}{PEFT}

\usepackage{etoolbox}
\makeatletter
\patchcmd{\maketitle}
 {\def\@makefnmark}
 {\def\@makefnmark{}\def\useless@macro}
 {}{}
\makeatother

\title{When does Parameter-Efficient Transfer Learning Work \\for Machine Translation?}

\author{Ahmet \"{U}st\"{u}n* \thanks{*~Both authors contributed the paper equally and the order is determined by coin flip.} \\
  University of Groningen\\
  \texttt{a.ustun@rug.nl} \\\And
  Asa Cooper Stickland* \\
  University of Edinburgh \\
  \texttt{a.cooper.stickland@ed.ac.uk} \\}

\begin{document}
\maketitle
\begin{abstract}

Parameter-efficient fine-tuning methods (PEFTs) offer the promise of adapting large pre-trained models while only tuning a small number of parameters. They have been shown to be competitive with full model fine-tuning for many downstream tasks. However, prior work indicates that PEFTs may not work as well for machine translation (MT), and there is no comprehensive study showing when PEFTs work for MT.
We conduct a comprehensive empirical study of PEFTs for MT, considering (1) various parameter budgets, (2) a diverse set of language-pairs, and (3) different pre-trained models. 
We find that `adapters', in which small feed-forward networks are added after every layer, are indeed on par with full model fine-tuning when the parameter budget corresponds to 10\% of total model parameters. Nevertheless, as the number of tuned parameters decreases, the performance of PEFTs decreases. The magnitude of this decrease depends on the language pair, with PEFTs particularly struggling for distantly related language-pairs. 
We find that using \peft{}s with a larger pre-trained model outperforms full fine-tuning with a smaller model, and for smaller training data sizes, \peft{}s outperform full fine-tuning for the same pre-trained model.\footnote{Our code and scripts for reproducing the experiments are available at \url{https://github.com/ahmetustun/fairseq}}
\end{abstract}

\section{Introduction}

There has been enormous progress on scaling up neural machine translation (NMT) in the recent years, resulting in `massively multilingual' models that are capable of translating across many languages \citep{bapna2022building}. Most successful applications rely on sequence-to-sequence pre-training that (1) leverages web-scale monolingual data with a masking objective to build a multilingual backbone (parent) model \citep{liu-etal-2020-multilingual,song2019mass}, or (2) directly targets a many-to-many NMT system by mining parallel corpora \citep{fan2020beyond}. 

Standard practice is to fine-tune every parameter of a particular a pre-trained model to specialize it to a language pair (or domain) of interest \citep{zoph-etal-2016-transfer, neubig2018rapid}. However, if we require specialization to many language pairs or domains, the storage and time costs of full fine-tuning may become prohibitive. Moreover, as models grow ever larger, more efficient methods become attractive.

As an alternative to full model fine-tuning, several parameter-efficient fine-tuning methods (\textbf{PEFTs}) have been proposed. Such methods only fine-tune a small number of parameters, reducing storage cost, and avoid calculating the gradients for every model parameter, reducing training time and memory cost. Examples include adapters \citep{houlsby2019parameter,bapna-firat-2019-simple} and prefix-tuning \citep{li-liang-2021-prefix}, which introduce a few extra parameters to fine-tune, keeping the pre-trained model fixed. Others like BitFit \citep{zaken2021bitfit} tune only the bias vectors of the backbone model and similarly \citet{gheini-etal-2021-cross} update only cross-attention layers. 

PEFTs can produce results that are competitive with full fine-tuning. For instance, adapters can match full fine-tuning performance on the GLUE benchmark using only 2-4\% additional parameters \citep{houlsby2019parameter}. However their potential for MT has not been fully explored. Prior studies indicate that PEFTs designed for classification tasks can fail for MT \cite{adapt}, and it is not known how source and target language characteristics affect PEFTs' performance.
  
In this work, we provide a comprehensive analysis of PEFTs for MT. For our analysis, we consider: (1) different pre-trained models which vary in size from 484 million to 1.2 billion total parameters, (2) several PEFTs, and (3) typographically and geographically diverse languages. Moreover, we vary the number of tuned parameters, resulting in different parameter `budgets', ranging from 0.03\% to 10\% of total model parameters. 
Our main research questions are:
\begin{enumerate}[leftmargin=*,labelindent=0cm, label=RQ\arabic*:, ref=RQ\arabic*]
    \item \label{RQ1}For a given parameter budget, which \peft{} works best? 
    \item \label{RQ2}How does language similarity affect the performance of PEFTs for different parameter `budgets'?
    \item \label{RQ3}How does (i) the pre-training objective, and (ii) the size of the parent model affect the performance of \peft{}s?
    \item \label{RQ4}Do PEFTs work better than fine-tuning for small dataset sizes?
\end{enumerate}
\paragraph{Key Findings} \textbf{1)} We found methods which introduce new parameters to a pre-trained model, namely adapters and prefix tuning, give us the best performance (\S~\ref{sec:compare-methods}). \ahmet{As we increase the number of new parameters, adapters retain good performance, while prefix-tuning falls behind.}
\textbf{2)} We found a large variation in PEFTs' performance across language pairs. Specifically, the distance between the source and target languages is negatively correlated with performance, especially for methods tuning the smallest number of parameters and methods tuning a subset of existing parameters (like bias terms or cross attention) (\S~\ref{sec:lang-related}). \textbf{3)} We observe that increasing model size, but keeping the same number of fine-tuned parameters, substantially increases MT performance (\S~\ref{sec:model-size}).
Finally, \textbf{4)} we observe that adapters perform better than full fine-tuning for small datasets, with the advantage for adapters increasing as dataset size gets smaller (\S~\ref{sec:data-size}).

\section{Background}
\label{sec:background}

This section briefly describes the two multilingual pre-trained models that we focus on in this work, namely mBART and M2M-100. 

\paragraph{Multilingual Denoising Pre-training}

Multilingual BART, mBART \cite{liu-etal-2020-multilingual}, is a sequence-to-sequence transformer model \cite{vaswani2017attention} that consists of an encoder and an autoregressive decoder. It is pre-trained with a \textit{denoising} objective, reconstructing a document from a noisy version. mBART uses span masking and sentence permutation to noise the original document. It consists of 12 encoder and 12 decoder layers, with hidden dimension of 1024 and 16 attention heads. mBART is trained entirely on monolingual data that includes multiple languages and it has a large multilingual vocabulary of 250k tokens. In our experiments, we use mBART-50 \cite{tang2020multilingual} which was pre-trained on 50 languages.

\paragraph{Many-to-Many Multilingual MT}

The M2M-100 model \cite{fan2020beyond} is a many-to-many multilingual translation system that is pre-trained on a large-scale parallel dataset for 100 languages and 100$\times$99 translation directions. This dataset is automatically constructed with a novel data mining method based on language similarities and back-translation. The model is trained in a many-to-many fashion, balancing languages using \textit{sinkhorn} temperature sampling. In our experiments, we use the base size M2M-100 with 484M parameters that consists of 12 encoder and 12 decoder layers, hidden dimension of 1024 and feedforward dimension of 4096. To study the effect of model size, we also use the medium size M2M-100 with 1.2B parameters, which has 24 encoder and 24 decoder layers, and feedforward dimension of 8192. 
Both models have a multilingual vocabulary of 128K unique tokens that are distributed across 100 languages with temperature sampling. 

\section{Parameter Efficient Fine-tuning Methods}
\label{sec:methods}
All of our experiments fall under the umbrella of specialising a pre-trained sequence-to-sequence transformer model for MT of a particular language pair, with source language $x$ and target language $y$. If the pre-training task was MT, and $x$ and $y$ were included, then a lower bound will be simply applying the pre-trained model without any changes. Conversely an upper bound is fine-tuning 100\% of the pre-trained model parameters (`full fine-tuning'). In between full fine-tuning and directly using the pre-trained model, we consider the following parameter-efficient fine-tuning methods (\peft{}s) in this work:

\paragraph{Adapter-tuning {\normalfont\cite{houlsby2019parameter}}}
`Adapter layers' are lightweight, learnable units inserted between transformer layers. They typically take the form of a feedforward network inserted as the final operation in a transformer layer. Formally, we follow the architecture introduced by \citet{bapna-firat-2019-simple} for MT:
\begin{equation}
\textrm{A}_{\ell}(\mathbf{h}^{\ell}) = W^T_{\text{u}} \cdot f(W^T_{\text{d}} \textrm{LN}(\mathbf{h}^{\ell}) + \mathbf{b}^{\ell}_{\text{d}})+ \mathbf{b}^{\ell}_{\text{u}},
\end{equation}

\noindent where an adapter module $\textrm{A}_{\ell}$ at layer $\ell$ consists of a layer-normalization LN of the input $h^{\ell}\in \mathcal{R}^d$, followed by a down-projection $W_{\text{d}}\in \mathcal{R}^{d\times b}$ with bottleneck dimension $b$, a non-linear function $f(\cdot)$ and an up projection $W_{\text{u}}\in \mathcal{R}^{b\times d}$. Finally, a residual connection with input $h^{\ell}$ is added to the output of the adapter: $\mathbf{h}^{\ell} \rightarrow \textrm{A}_{\ell}(\mathbf{h}^{\ell}) + \mathbf{h}^{\ell}$. We write `adapter-$b$' to mean adapters with bottleneck dimension $b$ throughout this work.

\paragraph{Prefix-tuning {\normalfont\cite{li-liang-2021-prefix}}} prepends a sequence of continuous task-specific vectors (`prefixes') to the model input, in analogy to natural language prompts (e.g.\ `translate this sentence:') which the transformer can attend to, but the prefix consists entirely of free parameters. For each transformer layer, the prefix is replaced with a new set of vectors, increasing expressiveness. Concretely, we replace token embeddings by \begin{equation}
 \label{eq:concat}
     E_{p} = \text{Concat}(V^0, E),
 \end{equation} 
with \(E \in \mathcal{R}^{L\times d}\) the original token embeddings packed into a matrix, \(V^0 \in \mathcal{R}^{p\times d}\) the prefix vectors, and $L$ the original sequence length, $p$ the prefix length and $d$ model dimension. Before transformer layer $\ell$ we additionally set the first $p$ hidden states to a new prefix vector, i.e.\ \(H^{\ell}[\mbox{:}p, :] = V^{\ell}\) with \(H \in \mathcal{R}^{(L+p)\times d}\) the hidden states and \(V^{\ell} \in \mathcal{R}^{p\times d}\).

\begin{figure*}[t]
  \begin{subfigure}[b]{0.488\textwidth}
    \includegraphics[width=\textwidth]{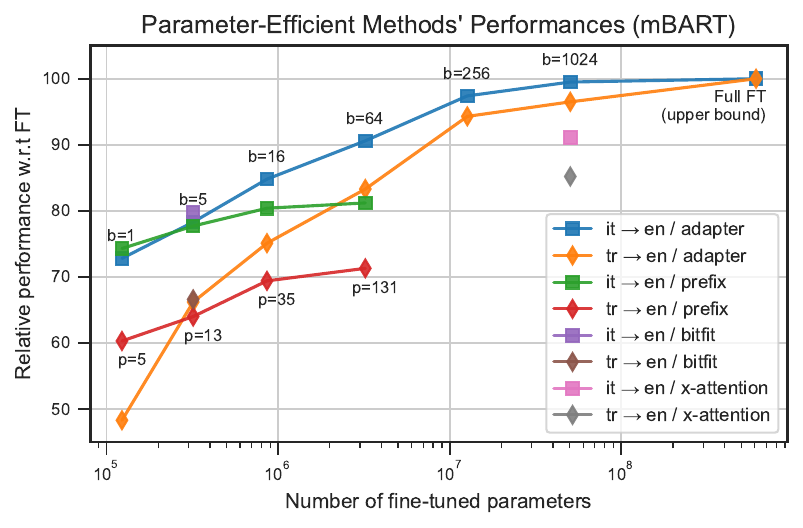}
  \end{subfigure} \vspace{0.5cm}
  \begin{subfigure}[b]{0.49\textwidth}
    \includegraphics[width=\textwidth]{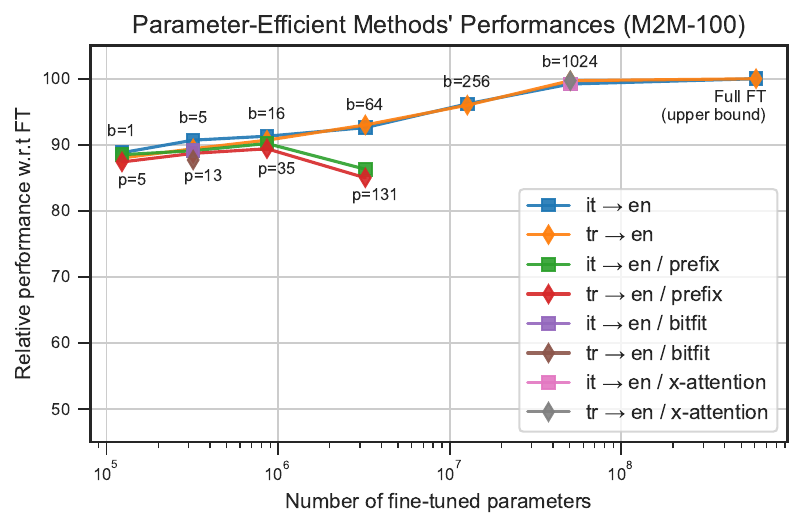}
  \end{subfigure} \vspace{-0.25cm}
    \caption{For increasing parameter budget, does prefix-tuning or adapters work best (\ref{RQ1})? We show relative MT performance over full fine-tuning vs.\ number of fine-tuned parameters for mBART and M2M-100. $b$ and $p$ refer to adapter bottleneck dimension and prefix length respectively. Due to the large effective sequence length, we limit prefix-tuning experiments.
    }
  \label{fig:adapter-size}
\end{figure*}

\paragraph{BitFit {\normalfont\cite{zaken2021bitfit}}} Bias term fine-tuning was introduced in the context of fine-tuning BERT for classification tasks, and consists of training only the bias terms and the task-specific classification layer. For MT we additionally fine-tune all decoder bias terms, and do not need the classification head. 
We introduce a simple improvement to BitFit, based on replacing redundant parameters with ones that increase expressiveness.
Note that BitFit fine-tunes bias parameters in layer-norm (LN) modules \cite{ba2016layer}, since the layer-norm contains the following affine transformation: 
\begin{equation}
 \label{eq:layernorm}
     \text{LN}^{\ell}_{\text{aff}}(\mathbf{z}^{\ell}) = \textcolor{black}{\gamma} \odot \mathbf{z}^{\ell} + \textcolor{black}{\beta}
 \end{equation} 
\noindent where $\mathbf{z}^{\ell}$ is the normalized input after a residual connection. \(\textcolor{black}{\gamma}, \textcolor{black}{\beta} \in \mathcal{R}^{d}\) are learnable weights and the bias parameters of the LN module. For the standard transformer model, the LN module is always followed by a matrix multiplication plus a bias term i.e.\ $W_{m}^{\ell} \cdot \text{LN}^{\ell}_{\text{aff}}(\mathbf{z}^{\ell}) +\textcolor{black}{b_{m}^{\ell}} = W_{m}^{\ell} \cdot \textcolor{black}{\gamma} \odot \mathbf{z}^{\ell}+ W_{m}^{\ell} \cdot\textcolor{black}{\beta} +\textcolor{black}{b_{m}^{\ell}}$. Notice the same space of functions is available by \textit{only} updating the $\textcolor{black}{b_{m}^{\ell}}$ term in $W_{m}^{\ell} \cdot \textcolor{black}{\beta} +\textcolor{black}{b_{m}^{\ell}}$.  We simply switch to updating $\textcolor{black}{\gamma}$ instead of $\textcolor{black}{\beta}$, i.e.\ unfreezing the LN weight and freezing the bias, in order to increase expressiveness (confirmed empirically in \S~\ref{sec:compare-methods}). We use this version of BitFit throughout this work unless stated otherwise.  

\paragraph{X-attention Tuning {\normalfont\cite{gheini-etal-2021-cross}}} refers to fine-tuning only cross-attention (X-attention) and corresponding layer-norm parameters located in each decoder layer of a transformer model. This method is based on the importance of cross-attention for MT. 

\begin{table}[t]
\small
\begin{tabular}{@{}lllll@{}}
\toprule
& \multicolumn{2}{c}{mBART} & \multicolumn{2}{c}{M2M-100} \\
& it$\rightarrow$en & tr$\rightarrow$en & it$\rightarrow$en & tr$\rightarrow$en \\ \midrule
Full FT   & 38.2 & 31.7 & 36.6 & 30.1 \\
\noalign{\smallskip} 
\hdashline 
\noalign{\smallskip}
X-attention & 34.8 & 27.0 & 36.1 & 29.2 \\ 
Adapter (b=1024) & \underline{38.0} & \underline{30.6} & 36.3 & \underline{30.0} \\ 
\noalign{\smallskip} 
\hdashline 
\noalign{\smallskip}
Prefix (p=13) & 29.7 & 20.3 & 32.7 & 26.7 \\ 
BitFit (LN-bias)   & 29.3 & 19.9 & 32.4 & 26.2 \\ 
BitFit (LN-weights)  & \underline{30.5} & 21.1 & 32.6 & 26.4 \\ 
Adapter (b=5) & 29.9 & \underline{21.9} & \underline{33.2} & \underline{26.9} \\
\noalign{\smallskip} 
\hdashline 
\noalign{\smallskip}
Prefix (p=5) & \underline{28.4} & \underline{19.1} & 32.4 & 26.3 \\
Adapter (b=1) & 27.8 & 15.3 & 32.5 & 26.5 \\
\bottomrule
\end{tabular}
\caption{For a given parameter budget, which method works best (\ref{RQ1})? BLEU scores for it$\rightarrow$en and tr$\rightarrow$en when different fine-tuning methods used for mBART and M2M-100. Each block consists of methods that update approximately the same number of parameters. We underline results which are significantly (p<0.05) best within a block w.r.t. \textit{paired bootstrap resampling}.
~chrF scores for these experiments are shown in Appendix~\ref{sec:appendix-results}.}
\label{tab:ft-methods-results}
\end{table}

\section{Experiments}
\label{sec:experiments}

\paragraph{Datasets} We conduct experiments with a selection of 12 typologically and geographically diverse languages, paired with English. In our experiments, we fine-tune the pre-trained model on only one language pair and translation direction at a time (e.g.\ Italian~$\rightarrow$~English). 
The parallel data for all languages is from TED talks in order to factor out the impact of the domain differences (except Finnish and Estonian which we only use for a separate control experiment). To pick these languages, we consider variation in language families and scripts. More details of the datasets are given in Appendix~\ref{sec:repro}.

\paragraph{Experimental Settings} 
We used mBART-50 \cite{liu-etal-2020-multilingual,tang2020multilingual} and M2M-100 \cite{fan2020beyond} as our multilingual pre-trained models, and all the languages we experiment with are included in their pre-training data. 
mBART needs to learn machine translation with parallel data, but M2M-100 can also be used without fine-tuning, since it is initially pre-trained for MT (see \S~\ref{sec:background}). 
We conduct experiments with both the base and the medium size M2M-100, to measure the impact of parent model size.  

For all 
fine-tuning methods, we fine-tuned models with a maximum learning rate of 1e-4 with 
2500 warm-up steps 
for 100K training updates. 
We picked the best model based on dev set perplexity. We used a maximum batch size of 1024 tokens for mBART and 600 tokens for M2M-100, with a gradient accumulation step (\textit{update-frequency}) of 2 for both models. All experiments are performed with the fairseq \cite{ott-etal-2019-fairseq} library. Additional details including dataset splits are in Appendix~\ref{sec:repro}.

We 
use BLEU scores to estimate MT quality, calculated from Sacrebleu\footnote{Sacrebleu signature (BLEU):\\nrefs:1|case:mixed|eff:no|tok:13a|smooth:exp|version:2.0.0 \\ For the significance test we used \texttt{--paired-bs} flag.} \cite{post-2018-call}. 
To compare fine-tuning methods across different languages, we often report \textbf{relative performance} with respect to full fine-tuning (FT) for each language by calculating the ratio of each method's BLEU score w.r.t.\ the full FT BLEU score.\footnote{BLEU scores for each direction are given in Appendix~\ref{sec:appendix-results}} 
On the recommendation of \citet{marie-etal-2021-scientific} we report chrF \cite{popovic-2015-chrf} in Appendix~\ref{sec:appendix-results} for each fine-tuning method.

\paragraph{Parameter Budget Selection} In order to fairly compare different methods, we selected a series of parameter `budgets', and adjusted the settings of each method such that they update the same number of parameters. To determine the parameter budgets, we used the number of trainable parameters for the cross-attention update and BitFit since these numbers are constant (Unlike adapters and prefix-tuning, where we have an adjustable bottleneck dimension). Additionally, when comparing adapters and prefix-tuning, we start from the parameter size of the smallest adapter where the bottleneck dimension is~1.\footnote{Each block corresponds to parameter budgets of approximately 50m, 320k, and 120k trainable parameters, representing \{X-attention, Adapter-1024\}, \{BitFit, Adapter-5, Prefix-13\}, and \{Adapter-1, Prefix-5\} respectively.}\looseness=-1

\begin{figure*}[t]
  \begin{subfigure}[b]{0.488\textwidth}
    \includegraphics[width=\textwidth]{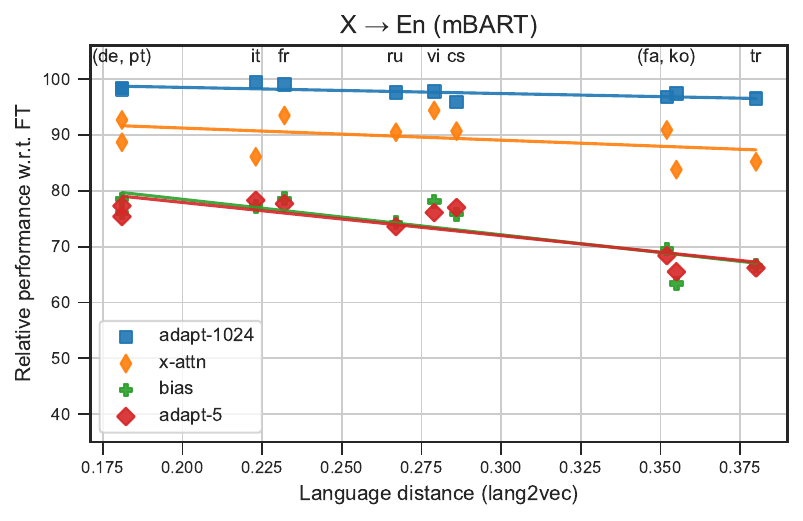}
  \end{subfigure} 
  \begin{subfigure}[b]{0.49\textwidth}
    \includegraphics[width=\textwidth]{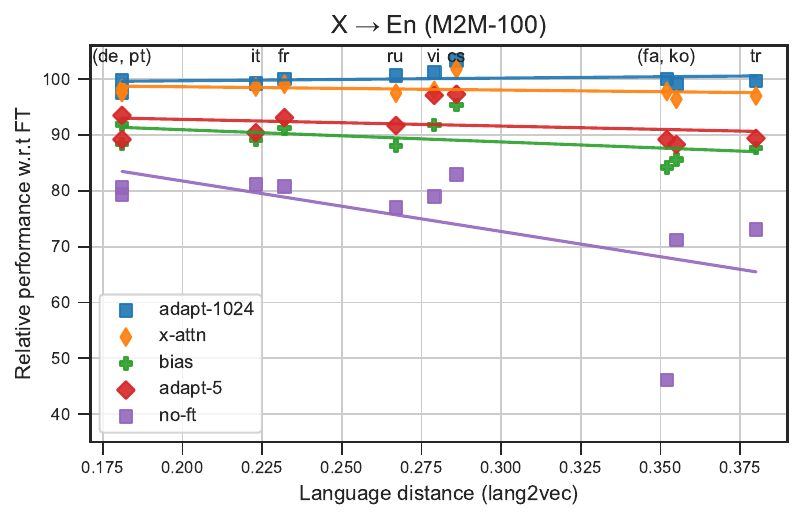}
  \end{subfigure}
  \caption{How does language similarity affect relative performance in x$\rightarrow$en with respect to full fine-tuning (\%) for PEFTs (\ref{RQ2})? Trend lines show the correlation between performance of PEFTs and language distance.}
  \label{fig:mbart-language-sim}
\end{figure*}

\begin{figure*}[h]
  \begin{subfigure}[b]{0.488\textwidth}
    \includegraphics[width=\textwidth]{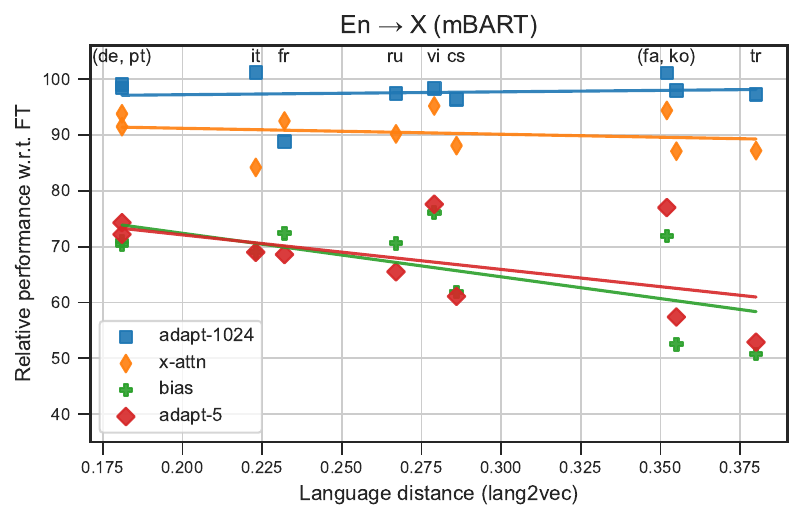}
  \end{subfigure} 
  \begin{subfigure}[b]{0.49\textwidth}
    \includegraphics[width=\textwidth]{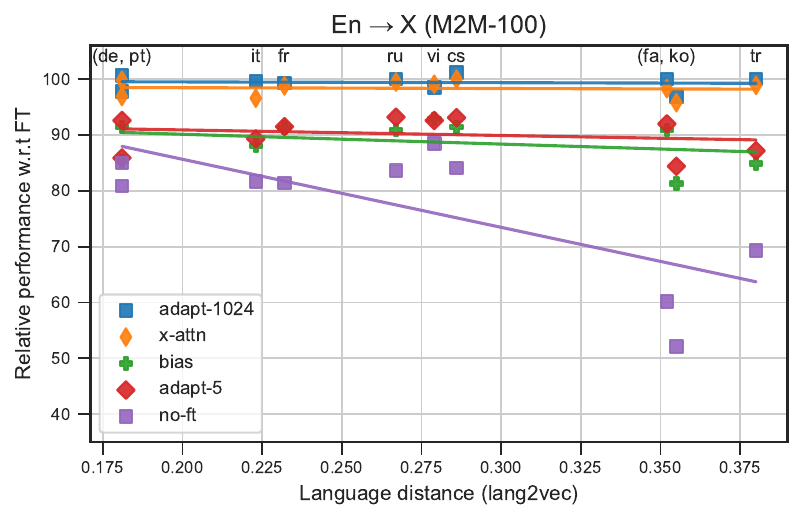}
  \end{subfigure}
 \caption{How does language similarity affect relative performance in en$\rightarrow$x with respect to full fine-tuning (\%) for PEFTs (\ref{RQ2})? Trend lines show the correlation between performance of PEFTs and language distance.}
  \label{fig:m2m-language-sim}
\end{figure*}

\section{Results and Discussion}
In this section, we first compare the performance of various \peft{}s on two language directions for different parameter budgets \S~\ref{sec:compare-methods}. We then select a subset of these methods to test on ten language directions, in order to evaluate the effect of language similarity on the performance of \peft{}s \S~\ref{sec:lang-related}. We use these results to explore the effect of parent model pre-training \S~\ref{sec:pretraining-type} and parent model size \S~\ref{sec:model-size}. We noticed that on the language directions with the smallest dataset size, adapter methods outperformed full fine-tuning, and therefore conducted control experiments showing that as dataset size decreases, adapters outperform full fine-tuning by a larger margin.

\subsection{RQ1: Comparing fine-tuning methods}
\label{sec:compare-methods}

Table~\ref{tab:ft-methods-results} shows the \textbf{performance} of \peft{}s in terms of BLEU score for 
it$\rightarrow$en and tr$\rightarrow$en.
In the table, each block (separated with a dashed line) consists of PEFTs with approximately the same number of updated parameters.
Adapters outperform other methods for almost all parameter budgets for both mBART and M2M-100, 
except the smallest budget of 120k updated parameters. In this block, prefix-tuning (prefix-5) performs better than adapters for mBART. However, when the fine-tuned parameter count increases, as shown in Figure \ref{fig:adapter-size}, prefix-tuning quickly falls behind adapters, confirming previous findings \cite{unified}. Furthermore, in terms of \textbf{training speed/memory cost}, prefix-tuning slows down training relative to adapters, and imposes a significant memory cost due to a large effective sequence length; see also Appendix~\ref{sec:prefix-details}.\footnote{Prefix-13 causes a 30\% slow-down in training speed relative to adapter-5.} 

As for the methods that fine-tune existing parameters, both BitFit and X-attention performs worse than adapters in most cases. Averaging across 10 language pairs, adapters still outperform BitFit for both parent models (Figure~\ref{fig:mbart-vs-m2m}). However, we confirm that our method of tuning layer norm weights rather than biases improves BitFit, see Table~\ref{tab:ft-methods-results}. 


\begin{figure*}[t]
  \begin{subfigure}[b]{0.255\textwidth}
    \includegraphics[width=\textwidth]{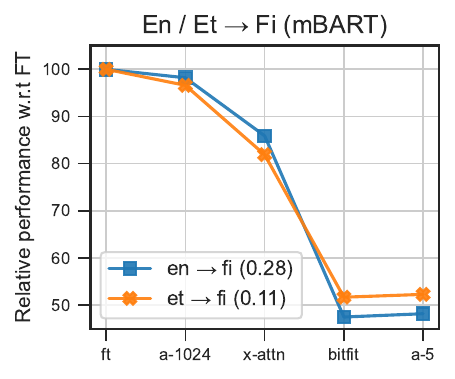}
  \end{subfigure} 
  \begin{subfigure}[b]{0.235\textwidth}
    \includegraphics[width=\textwidth]{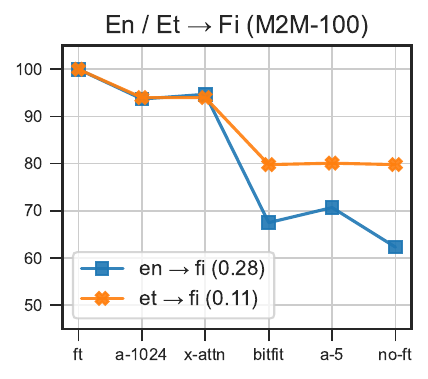}
  \end{subfigure}
  \begin{subfigure}[b]{0.24\textwidth}
    \includegraphics[width=\textwidth]{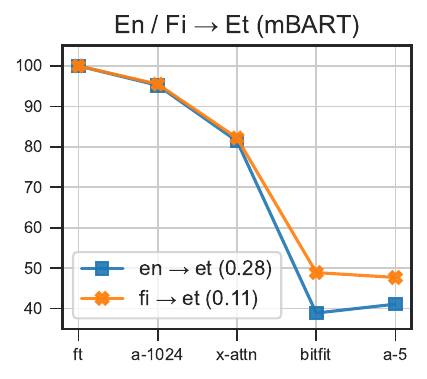}
  \end{subfigure} 
  \begin{subfigure}[b]{0.235\textwidth}
    \includegraphics[width=\textwidth]{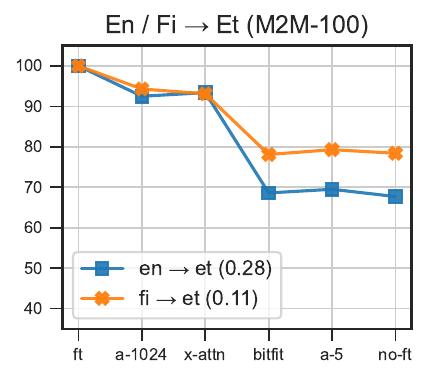}
  \end{subfigure}
  \caption{Decrease in relative performance (\%) over full fine-tuning as the number of updated parameters decreases for translating into Finnish and Estonian with different source language (en, et, fi). \texttt{Lang2vec} distances for en$\leftrightarrow$fi, en$\leftrightarrow$et, and fi$\leftrightarrow$et are 0.28, 0.28 and 0.11 respectively.}
  \label{fig:into-fi/et}
\end{figure*}

\subsection{RQ2: Impact of language relatedness}
\label{sec:lang-related}

In order to evaluate how language similarity between translation pairs affects the performance of different PEFTs, we extend our experiments to 10 languages paired with English (x$\rightarrow$en, en$\rightarrow$x), representing a diverse set of linguistic typology. Figure~\ref{fig:mbart-language-sim} and \ref{fig:m2m-language-sim} show performance w.r.t. full fine-tuning, for both mBART and M2M.

We found that similarity between source and target languages impacts the performance of PEFTs, with distantly related languages (e.g.\ English and Korean) leading to lower performance for the methods with a small number of updated parameters such as BitFit and adapter-5. And so when translating between distantly related languages, we need to tune more parameters to match full fine-tuning and get the most out of the parent model.

More concretely, relative performance w.r.t.\ full FT is negatively correlated with language distance measured by \texttt{lang2vec}\footnote{lang2vec is a python package based on the URIEL typology database \citep{littell2017uriel}. For language distance, we compute the cosine distance between typological feature vectors of languages that consists of syntactic, phonological and inventory features (289 features in total).}. These correlations are stronger for mBART than M2M. Methods which tune existing parameters (X-attention and BitFit) and M2M with no fine-tuning show higher correlation than adapters with similar parameter budgets; see Table~\ref{tab:correlation}. One explanation is that adding parameters, and therefore increasing model capacity with adapters is beneficial for overcoming the difficulty of translating distant languages. 

We provide correlation results with more fine-grained measures of language distance, namely syntactical, phonological and geographical distances in Appendix~\ref{sec:extracorrel}. For the first two distances, we observe a similar trend: as the distance between source and target language increases, BitFit and small adapters do not perform as well (the negative correlation is stronger). Generally the syntactic features produced a larger negative correlation than the phonological features, with the exception of M2M plus PEFTs for en$\rightarrow$x. However, in terms of geographic distance, we do not observe a particular trend. 

\begin{table}[t]
\small
\begin{tabular}{@{}lllll@{}}
\toprule
& \multicolumn{2}{c}{mBART} & \multicolumn{2}{c}{M2M-100} \\
&x$\rightarrow$en & en$\rightarrow$x & x$\rightarrow$en & en$\rightarrow$x \\ \midrule
Adapter (b=1024) & \textit{-0.43} & \textit{0.11} & \textit{0.23} & \textit{-0.07}\\ 
X-attention & -0.69& \textit{-0.21} &\textit{-0.28} & \textit{-0.07}\\ 
\noalign{\smallskip} 
\hdashline 
\noalign{\smallskip}
Adapter (b=5) & -0.85 & \textit{-0.53} & \textit{-0.26} & \textit{-0.22}\\
BitFit & -0.84 & -0.64 & \textit{-0.47} & \textit{-0.33}\\ 
\noalign{\smallskip} 
\hdashline 
\noalign{\smallskip}
No FT & ~~~- & ~~~- & -0.60 &-0.72\\
\bottomrule
\end{tabular}
\caption{Pearson correlation coefficients between relative performance w.r.t.\ fine-tuning and language distance. \ahmet{Negative correlation means that relative performance tends to decrease as the distance between source and target language increases}. Numbers in italics are not statistically significant (p=0.05).}
\label{tab:correlation}
\end{table}

To investigate whether our findings extend beyond English-centric settings, we designed another set of experiments. We picked 3 languages from MultiParaCrawl, Finnish, Estonian and English, where Finnish and Estonian are from the same language family and typologically similar. 
We measure translation performance into Finish from Estonian and English, for different fine-tuning methods, and similarly for translation into Estonian. Figure~\ref{fig:into-fi/et} shows results
for both mBART and M2M-100. 

As shown in the first two plots, 
when translating into Finnish, Estonian as the source language gives an advantage over English for BitFit and adapter-5 (This advantage is higher in M2M-100 
than mBART). 
Likewise, for translation into Estonian, 
as the number of trainable parameters decreases, relative MT performance drops less when Finnish is the source language compared to English, for both parent models. Thus, when the source and target languages are typologically similar, PEFTs make better use of the parent model. 

\subsection{RQ3: Impact of parent model}
\label{sec:pretraining-type}

\begin{figure}[t]
    \centering
    \hspace{-0.5cm}
    \includegraphics[scale=0.65]{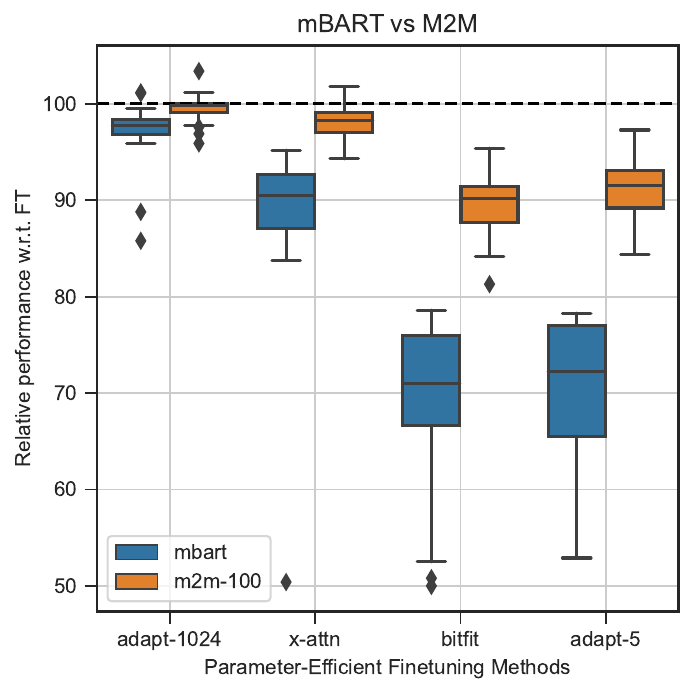}
    \caption{How does the type of parent model (different pre-training objective) affect performance (\ref{RQ3})? Statistics of relative performance w.r.t.\ full fine-tuning (\%) for all languages (x$\leftrightarrow$e) when the model is initialized with mBART or M2M-100. The dashed line refers to full fine-tuning performance.}
  \label{fig:mbart-vs-m2m}
\end{figure}

\paragraph{Pre-training Objective} Figure~\ref{fig:mbart-vs-m2m} shows the overall performances for PETFs aggregated over all languages (x$\leftrightarrow$en) when the model is initialized with mBART or M2M-100. In general, 
PEFTs for M2M-100 provides higher \textit{relative} performance than mBART (Fig.~\ref{fig:mbart-vs-m2m}). This difference is larger when the number of trainable parameters is small (BitFit and adapter-5). 
While M2M-100 is pre-trained for MT with parallel data, mBART is pre-trained with a (monolingual\footnote{Although mBART-50 pre-trained on 50 languages, the pre-training objective does not use any cross-lingual signal.}) denoising objective. Thus, more parameters are required at fine-tuning time to `learn' the MT task for mBART. 
Finally, we note mBART results have a higher variance than M2M-100 (see Fig.~\ref{fig:mbart-vs-m2m}), due to the higher negative correlation with language distance.

\paragraph{Model Size}
\label{sec:model-size}
We investigate how parent model size affects the performance of fine-tuning methods, comparing M2M-100's base model (484M) 
to its medium model (1.2B). 
Table~\ref{tab:size} shows the average performance of full fine-tuning and small-size adapters
\ahmet{corresponding to approximately 300K new parameters\footnote{Both adapter-5 in the base model and adapter-2 in the medium model correspond to roughly the same number of trainable parameters (0.07\% of 484M and 0.03\% of 1.2B total parameters).}.} No fine-tuning (no FT) results are also shown, representing \textit{lower} bounds.

Predictably, the medium model outperforms the base model across all fine-tuning methods. The magnitude of this improvement is larger when translating into English (\textit{x$\rightarrow$en}) vs.\ \textit{x$\rightarrow$en}, and the increase for small adapters is larger than for other methods.
When translating into English, small adapters with the medium model outperform \textit{full fine-tuning} of the base model for most languages despite tuning only 0.03\% of its parent model parameters.
For \textit{en$\rightarrow$x}, small adapters are still competitive with full fine-tuning of the base model with almost the same average performance.
But for distantly related languages to English (Farsi, Korean and Turkish), adapters' (1.2B) performance falls behind full fine-tuning of the base model. 

\begin{table}[t]
\small
\begin{tabular}{@{}lccc@{}}
\toprule
& \multicolumn{2}{c}{Model} &  \\
& Base (418m) & Med. (1.2b) & $\Delta$ BLEU\\
\midrule
$en\rightarrow$x & & & \\
\noalign{\smallskip} 
No FT & 21.9 & 24.3 & 2.4 \\
Small adapter & 24.8 & 27.4 & 2.6 \\
Full FT & 27.4 & 28.4 & 1.0 \\
\noalign{\smallskip} 
\hdashline 
\noalign{\smallskip}
$x\rightarrow$en & & & \\
No FT & 26.1 & 28.5 & 2.4 \\
Small adapter & 31.7 & 35.3 & 3.6 \\
Full FT & 34.4 & 36.6 & 2.2 \\
\bottomrule
\end{tabular}
\caption{How does parent model size affect performance (\ref{RQ3})? Average BLEU score across 10 languages for the base (484m parameters) and medium (1.2 billion parameters) M2M parent models, when tuning all parameters (`full FT'), when tuning small adapters, and when tuning no parameters (`no FT'). We also show the increase in BLEU when moving from the base to the medium model. See Appendix~\ref{sec:appendix-results} for individual results.}
\label{tab:size}
\end{table}

When it is used without \textit{any} parameter updates (`no FT'), the medium model (while outperforming the base model for no FT) is not competitive with small size adapters for the base model, in either direction (\textit{x$\leftrightarrow$en}). Furthermore, relative performance w.r.t. full fine-tuning is still negatively correlated with language distance (see Appendix Table~\ref{tab:billion-correl}). 
Therefore, even at large scales, parameter efficient fine-tuning is useful, taking MT performance to the \textit{upper} bound of a smaller model. 

\begin{figure*}[t]
  \begin{subfigure}[b]{0.3\textwidth}
    \includegraphics[width=\textwidth]{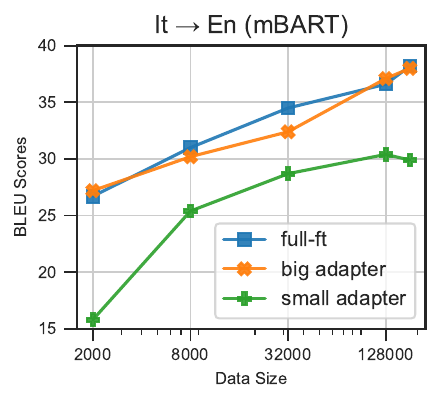}

  \end{subfigure} 
  \begin{subfigure}[b]{0.3\textwidth}
    \includegraphics[width=\textwidth]{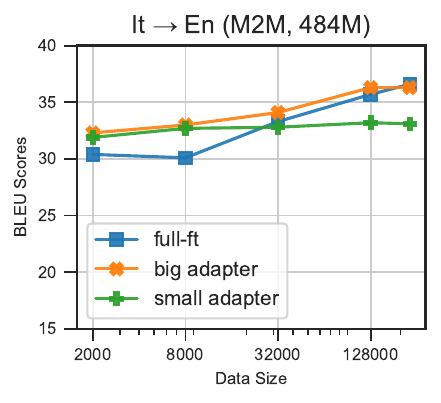}

  \end{subfigure}
  \begin{subfigure}[b]{0.3\textwidth}
    \includegraphics[width=\textwidth]{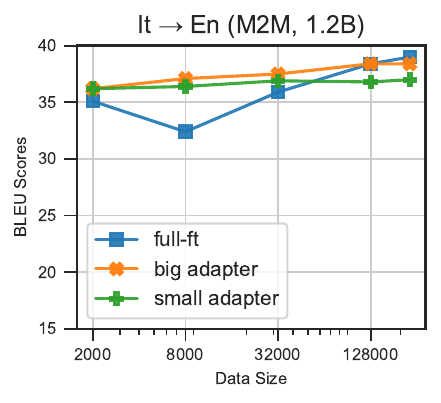}

  \end{subfigure} 
  \begin{subfigure}[b]{0.3\textwidth}
    \includegraphics[width=\textwidth]{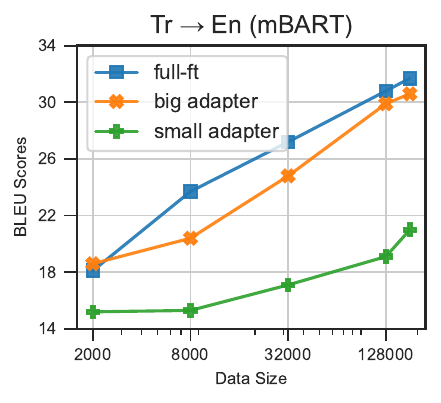}

  \end{subfigure}
  \begin{subfigure}[b]{0.3\textwidth}
    \includegraphics[width=\textwidth]{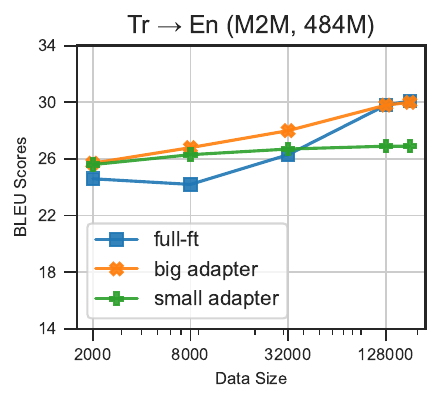}

  \end{subfigure}
  \begin{subfigure}[b]{0.3\textwidth}
    \includegraphics[width=\textwidth]{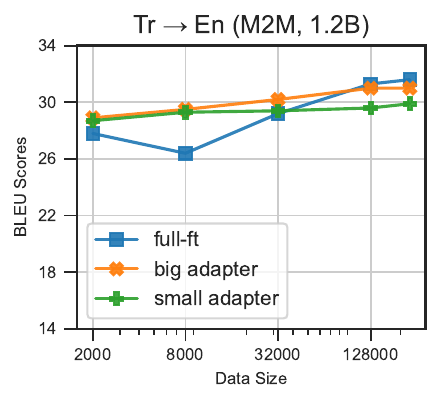}

  \end{subfigure}
  \caption{How does the amount of parallel data affect fine-tuning performance (\ref{RQ4})? \ahmet{BLEU scores for various subsets of the full training data for Italian to English and Turkish to English, with base (484m parameters) and medium (1.2 billion parameters) M2M and mBART parent models}. Big and small adapters have $\approx$50m and $\approx$300k parameters respectively for all models. 
  }
  \label{fig:dataset-size}
\end{figure*}

\subsection{RQ4: Impact of fine-tuning dataset size}
\label{sec:data-size}
We noticed that for the datasets with the smallest amount of training data (Vietnamese and Czech), \peft{}s outperformed full fine-tuning (see Appendix~\ref{sec:appendix-results}).
We therefore designed a control experiment to test for the effect of the training data size on \peft{}' performance, taking a random subset of sizes 2000, 8000, 32000 and 128000 training examples for Italian to English and Turkish to English. We then evaluated full fine-tuning, large adapters ($\approx$50m parameters) and small adapters ($\approx$300k parameters) on each dataset; see Figure~\ref{fig:dataset-size}.

For all models, at the smallest dataset size, large adapters outperformed full fine-tuning, and for M2M full fine-tuning only catches up at 128k examples. For mBART, small adapters lag far behind, indicating they do not provide enough capacity to `learn' the MT task. For M2M however, small adapters are on a par with larger ones for small dataset sizes, but fall behind as dataset size increases. Again, we believe this is because more capacity is needed to get the most out of larger datasets.

\citet{pet} explore the effect of fine-tuning dataset size for RoBERTa fine-tuned on English NLU tasks, finding \peft{}s outperform full fine-tuning for dataset size <1000. Interestingly, for mBART, similarly small dataset sizes are required for outperforming full fine-tuning. However, for M2M, we see adapters outperforming up until dataset sizes of $\approx128$k. 
Perhaps the `gap' between RoBERTa's masked language model pre-training objective and the fine-tuning objective is similar to the gap between mBART's pre-training objective and MT, whereas since M2M is pre-trained for MT, leaving the base model unchanged is viable up to larger fine-tuning dataset sizes. We leave further exploration of this to future work. Finally, we observe that full fine-tuning always converges in fewer iterations than the adapter methods, in a result similar to that of \citet{pet}.

\section{Related Work}

\peft{}s have been widely used for fine-tuning Transformer models to new tasks, domains or languages.
Adapters \citep{houlsby2019parameter} have been used in multi-task learning \cite{stickland2019bert,pfeiffer-etal-2021-adapterfusion,karimi-mahabadi-etal-2021-parameter}, cross-lingual transfer \cite{ustun-etal-2020-udapter,pfeiffer2020mad} and multilingual NMT \cite{bapna-firat-2019-simple,philip-etal-2020-monolingual,cooper-stickland-etal-2021-recipes, ustun-etal-2021-multilingual}. 
Prefix-tuning \cite{li-liang-2021-prefix} and Prompt-tuning \cite{lester-etal-2021-power, qin-eisner-2021-learning} (i.e.\ only using soft prompt tokens without prefix vectors in each layer), have a natural interpretation in terms of virtual tokens. They can be used as task embeddings for inter-task transferability \cite{vu2021spot}. 
LoRA \cite{lora} injects trainable low-rank matrices into query and value projection matrices of each transformer layer.
\citet{unified} present a unified framework that integrates the above methods. 


Some of these methods have been compared in a controlled setting for English classification tasks \cite{pet} or only a single language pair (English and Romanian) for MT \cite{unified}. 
\citet{pet} test \peft{}s for various English classification tasks and observe that on the tasks with the smallest dataset sizes, \peft{}s outperform fine-tuning, but they do not conduct a control experiment varying dataset size and parent model for a single task as we do.

Aspects of efficiency and scale in MT in terms of inference cost \cite{berard-etal-2021-efficient}, vocabulary size \cite{gowda-may-2020-finding} data \cite{gordon-etal-2021-data}, model size \cite{gordon-etal-2021-data,arivazhagan} and number of languages \cite{arivazhagan} have been explored. Other work aims to improve full FT for domain adaptation by mixing in different data \cite{chu-etal-2017-empirical}, regularisation \cite{miceli-barone-etal-2017-regularization} or many other methods \cite{chu-wang-2018-survey, survey}. However, none of these works study \peft{}s for MT, and we aim to fill this gap. 

\section{Conclusion}

Do \peft{}s work for MT? \ahmet{We found that the answer depends on multiple factors: the particular method, the backbone model, the number of tuned parameters and the fine-tuning language pair.} Adapters usually have the highest performance out of all \peft{}s (\S~\ref{sec:compare-methods}), although for the smallest parameter budgets we consider, prefix tuning outperforms adapters for mBART.
For large parameter budgets ($\approx$50m parameters) adapters almost recover full fine-tuning performance, and even for lower budgets, if the pre-training task was MT, i.e.\ M2M-100, adapters can recover $>$90\% of full FT performance. However \peft{}s only \textbf{outperform} full FT for smaller dataset sizes (\S~\ref{sec:data-size}), less than around $\approx$2k examples for mBART and $\approx$128k for M2M. Future work could explore in detail how the difference between pre-training objective and fine-tuning task affects this phenomenon.

Using \peft{} with a larger model (M2M-100 medium size) can outperform full FT of a smaller model (M2M-100 base size). However when translating in the  en$\rightarrow$x direction where $x$ is distantly related to English e.g.\ Korean, full FT is superior (\S~\ref{sec:model-size}). More generally, distantly related language pairs require more parameters to be tuned to get close to full FT, for all methods (\S~\ref{sec:lang-related}). 


\section{Limitations}

Firstly, in this work we do not cover all parameter-efficient fine-tuning methods (or variations on those that we do analyse) such as LoRA \citep{lora}, or mix-and-match adapters \citep{he2021towards}. In order to make our analysis compact and clear we center our comparison around simple adapters and prefix-tuning, together with BitFit and updating cross-attention.  
Secondly, our experiments only cover models with up to around 1 billion parameters due to compute limitations, which does not include the largest models available, such as the 11 billion parameter M2M or mT5 \cite{xue-etal-2021-mt5} models similar to the encoder-decoder models we use in this paper, or much larger autoregressive (and trained largely on English data) language models e.g.\ \citet{palm}.
Thirdly, although we attempted to cover a diverse set of languages, we did not explore truly low resource languages, and those not included in the pre-training data of our models (introducing another confounding factor for our language distance analysis), where one would expect even larger performance gaps for \peft{}s. However, we do imitate a very-low resource setup by limiting training data size (Section \S~\ref{sec:data-size}).
Furthermore, although we attempt to look into PEFTs' performances across languages w.r.t. different distance metrics such as syntax, phonology and geography (Appendix \S~\ref{sec:extracorrel}), 
more analysis in terms of fine-grained attributes such as word order or morphology are not provided in our analysis, which we leave for future work.  
Finally, we use automatic/string-based quality metrics, BLEU and chrF++ \cite{popovic-2017-chrf}, rather than pre-trained/neural quality metrics, with the latter often better correlated with human judgements \cite{kocmi-etal-2021-ship}.

\section*{Acknowledgements}
We would like to thank Iain Murray, Arianna Bisazza, Gosse Bouma, and Gertjan van Noord for valuable comments on a draft of this paper. We also would like to thank the Center for Information Technology of the University of Groningen for providing access to the Peregrine HPC cluster. Asa Cooper Stickland was supported in part by the EPSRC Centre for Doctoral Training in Data Science, funded by the UK Engineering and Physical Sciences Research Council (grant EP/L016427/1) and the University of Edinburgh.

\bibliography{anthology,custom}
\bibliographystyle{acl_natbib}


\appendix
\section{Reproducibility Report}
\label{sec:repro}

\paragraph{Datasets}

\begin{table*}[t]
\small
\begin{tabular}{@{}llllll@{}}
\toprule
\multirow{2}{*}{Language}&Language & Dataset & Train & Dev & Test \\ 
& family & source & size (k) & size (k) & size (k) \\ \midrule
Czech \textbf{(cs)}   & Slavic         & TED              & 103    & 3.5 & 3.8      \\
French \textbf{(fr)}   & Romance        & TED             & 192    & 4.3 & 4.9      \\
Korean \textbf{(ko)}   & Korean         & TED             & 205    & 4.4& 5.6      \\
Russian \textbf{(ru)}   & Slavic         & TED             & 208   & 4.8& 5.5       \\
Italian \textbf{(it)}   & Romance        & TED              & 231  & 0.9 & 1.6        \\
Portuguese \textbf{(pt)}   & Romance        & TED\textsuperscript{\textdagger}              & 184  & 4 & 4.9          \\
Turkish \textbf{(tr)}   & Turkic         & TED             & 182   & 4& 5         \\
Vietnamese \textbf{(vi)}   & Austri-Asiatic  & TED\textsuperscript{\textdagger}            & 133   &1.6 &1.3       \\
German \textbf{(de)}   & Germanic       & TED\textsuperscript{\textdagger}              & 206     &0.9 & 1.6     \\
Farsi \textbf{(fa)}   & Iranian        & TED             & 150    &3.9 & 4.5      \\
Finnish* \textbf{(fi)}   & Finnic         & mParacrawl           &   200 & 3         & 3 \\
Estonian* \textbf{(et)}   & Finnic         & mParacrawl              &  200  & 3         & 3\\ \bottomrule
\end{tabular}
\caption{Details of dataset that is used in our experiments. We gather language pairs (x$\leftrightarrow$en) from TED \cite{qi-etal-2018-pre} and IWSLT\textsuperscript{\textdagger} \cite{cettoloEtAl:EAMT2012} that are both compiled from TED talks.  
`*' indicates a set of separate controlled experiments where we randomly sampled 200k parallel sentences from MultiParacrawl (mParaCrawl) for corresponding language pairs.}
\label{tab:language-splits}
\end{table*}

All datasets that are used in our experimetns are publicly available. We used TED talks \cite{qi-etal-2018-pre} for (cs, fr, ko, ru, pt, tr, fa)$\leftrightarrow$en, IWSLT15 and IWSTL17 \cite{cettoloEtAl:EAMT2012} for vi$\leftrightarrow$en and (it, de)$\leftrightarrow$en respectively, IITB \cite{kunchukuttan-etal-2018-iit} for hi$\leftrightarrow$en. Finally, for (en, et, fi) experiments, we randomly sampled 200k parallel sentences for each language-pair from MultiParacrawl by using OPUS \cite{TIEDEMANN12.463}. Sizes of train, dev and test splits are given in Table \ref{tab:language-splits}. All datasets have licenses allowing non-commercial use. 

\paragraph{Pre-trained models and Hyper-parameters}
We used mBART \cite{liu-etal-2020-multilingual} that is extended to 50 languages \cite{tang2020multilingual}. For M2M-100 \cite{fan2020beyond}, we used base- and medium-size models that consist of 484M and 1.2B parameters. 

For all experiments we used the hyper-parameters that are reported by \citet{liu-etal-2020-multilingual} except learning rate. For the learning rate, we follow \citet{ustun-etal-2021-multilingual} and used maximum of 1e-4 with polynomial learning rate decay, based on their adapter-tuning experiments. We fine-tune models by using 0.3 dropout, 0.2 label smoothing, 2500 warm-up steps for 100K training updates with an early-stopping patience of 10 epochs. We used a maximum batch size of 1024 tokens for mBART and 600 tokens for M2M-100, with a gradient accumulation step (\textit{update-frequency}) of 2 for both models. For full fine-tuning (and not other methods) with the 1.2 billion size M2M model we use the Adafactor optimizer \cite{adafactor} in order to save memory (and use learning rate 5e-5), and otherwise use the Adam optimizer \cite{kingma2014adam}. We report the result of a single random seed/training run throughout this work whenever we list BLEU scores. All parameter-efficient fine-tuning methods are implemented on top of the Fairseq framework \cite{ott-etal-2019-fairseq}. We will share our code and scripts to reproduce all experiments.  

\paragraph{Computing Budget and Infrastructure}
All the experiments are conducted using Tesla V100 GPUs with mixed precision (\texttt{fp16}). Parameters that are fine-tuned for each model are reported in the experiments section (\S~\ref{sec:experiments}). Each individual experiment took 3-10 hours on one GPU depending on the fine-tuning method and the language-pair. 
\section{Prefix-tuning Details}
\label{sec:prefix-details}

There is relationship between memory cost and training time for prefix-tuning: including virtual tokens in a sentence will increase the effective length of that sentence, and we can either impose additional memory cost for the virtual tokens, or we can reduce the total number of `real' i.e. natural language as opposed to virtual tokens in each batch. With the latter method we avoid a large memory cost, however the time taken to iterate through a given number of training examples will be longer, since the number of real tokens per batch will be decreased, increasing training time. We use the latter (decreased `real' tokens) method.

Finally we note that inference speed will decrease as we increase the number of virtual tokens, since the decoder attention needs to attend to virtual tokens, i.e.\ when decoding token $n$ it will attend to $n -1 + p$ previous tokens for prefix length $p$. 

\section{Additional Results and Metrics}
\label{sec:appendix-results}

\begin{table}[t]
\small
\begin{tabular}{@{}lllll@{}}
\toprule
& \multicolumn{2}{c}{mBART} & \multicolumn{2}{c}{M2M-100} \\
&x$\rightarrow$en & en$\rightarrow$x & x$\rightarrow$en & en$\rightarrow$x \\ \midrule
\textbf{\textit{syntax}} \\
\noalign{\smallskip} 
\hdashline 
\noalign{\smallskip}
Adapter (b=1024) & -0.65 & \textit{0.17} & \textit{0.14} & \textit{-0.11}\\ 
X-attention & \textit{-0.36} &\textit{-0.0} & \textit{-0.34} & \textit{-0.06} \\ 
\noalign{\smallskip} 
\hdashline 
\noalign{\smallskip}
Adapter (b=5) & -0.85 & \textit{-0.33} & \textit{-0.26} & \textit{-0.21}\\
BitFit & -0.80 & -0.53 & \textit{-0.49} & \textit{-0.28}\\ 
\noalign{\smallskip} 
\hdashline 
\noalign{\smallskip}
No FT & ~~~- & ~~~- & -0.69 &-0.74\\
\midrule
\textbf{\textit{phonology}} \\
\noalign{\smallskip} 
\hdashline 
\noalign{\smallskip}
Adapter (b=1024) & \textit{-0.01} & \textit{-0.24} & \textit{0.01} & -0.71\\ 
X-attention & \textit{0.14} & \textit{0.16}& \textit{-0.35} & -0.47 \\ 
\noalign{\smallskip} 
\hdashline 
\noalign{\smallskip}
Adapter (b=5) & -0.45 & \textit{-0.06} & \textit{-0.28} & -0.42\\
BitFit & \textit{-0.42} & \textit{-0.13} & \textit{-0.42} & \textit{-0.34}\\ 
\noalign{\smallskip} 
\hdashline 
\noalign{\smallskip}
No FT & ~~~- & ~~~- & \textit{-0.36} &-0.54\\
\midrule
\textbf{\textit{geography}} \\
\noalign{\smallskip} 
\hdashline 
\noalign{\smallskip}
Adapter (b=1024) & \textit{-0.18} & \textit{0.21} & \textit{0.07} & -0.58\\ 
X-attention & \textit{-0.02} & \textit{0.23} & \textit{0.21} & \textit{-0.24} \\ 
\noalign{\smallskip} 
\hdashline 
\noalign{\smallskip}
Adapter (b=5) & \textit{-0.42} & \textit{0.09} & \textit{0.05} & \textit{-0.13}\\
BitFit & \textit{0.09} & \textit{-0.06} & \textit{-0.29} & \textit{-0.13}\\ 
\noalign{\smallskip} 
\hdashline 
\noalign{\smallskip}
No FT & ~~~- & ~~~- & \textit{-0.28} &\textit{-0.36}\\
\bottomrule
\end{tabular}
\caption{Pearson correlation coefficients between relative performance w.r.t.\ fine-tuning and syntactic, phonological and geographical language distances. Negative correlation means that relative performance tends to decrease as the distance between source and target language increases. Numbers in italics are not statistically significant (p=0.05).}
\label{tab:correlation}
\end{table}

\begin{table}[t]
\begin{tabular}{c@{\hskip 0.1in}cc}
\toprule
\noalign{\smallskip}
& \textit{\textbf{en}$\rightarrow$\textbf{x}} & \textit{\textbf{x}$\rightarrow$\textbf{en}} \\
\noalign{\smallskip}
Adapter (b=2) & -0.75 & \textit{-0.39} \\
No fine-tuning & -0.75 & \textit{-0.55} \\
\bottomrule
\end{tabular}
\caption{Correlation coefficients between language distance and relative performance for the 1.2 billion size M2M model; see also Table~\ref{tab:correlation}. Numbers in italics are not statistically significant (p=0.05).}
\label{tab:billion-correl}
\end{table}

Table~\ref{tab:ft-methods-results-chrF} shows chrF \cite{popovic-2017-chrf} scores\footnote{Sacrebleu signature (chrF2++):\\nrefs:1|case:mixed|eff:yes|nc:6|nw:2|space:no|version:2.0.0} for the experiments comparing different \peft{}s on it$\rightarrow$en and tr$\rightarrow$en (Table~\ref{tab:ft-methods-results}). These results confirms that the trends discussed in Section \ref{sec:experiments} are the same regardless of metric used for MT quality.

\begin{table}[t]
\small
\begin{tabular}{@{}lllll@{}}
\toprule
& \multicolumn{2}{c}{mBART} & \multicolumn{2}{c}{M2M-100} \\
& it$\rightarrow$en & tr$\rightarrow$en & it$\rightarrow$en & tr$\rightarrow$en \\ \midrule
Full FT             & 59.4 & 53.3 &   58.2   &  52.6 \\
\noalign{\smallskip} 
\hdashline 
\noalign{\smallskip}
X-attention         & 56.6 & 48.9 &     57.7 & 51.6     \\
Adapter (b=1024)    & \underline{59.2} & \underline{52.3} & 57.8 & \underline{52.2} \\
\noalign{\smallskip} 
\hdashline 
\noalign{\smallskip}
Prefix (p=13)       & 52.4 & 42.8 & 55.3 & 49.7 \\
BitFit (LN-bias)    & 51.8 & 41.7 & 55.0   & 49.3 \\
BitFit (LN-weights) & \underline{52.7} & 42.8 & 55.1     & 49.5     \\
Adapter (b=5)       & 52.4 & \underline{44.3} & \underline{55.5} & \underline{49.9} \\
\noalign{\smallskip} 
\hdashline 
\noalign{\smallskip}
Prefix (p=5)        & \underline{51.4} & \underline{41.4} & 54.9 & 49.5 \\
Adapter (b=1)       & 50.5 & 36.5 & 55.0   & 49.5 \\
\bottomrule
\end{tabular}
\caption{chrF scores for it$\rightarrow$en and tr$\rightarrow$en when different fine-tuning methods used for mBART and M2M-100. Each block represents same ratio of updated parameters. We underline results when a model is the significantly best within a block w.r.t. \textit{paired bootstrap resampling} test.}
\label{tab:ft-methods-results-chrF}
\end{table}

In Tables~\ref{tab:all-m2m}, \ref{tab:all-mBART} and \ref{tab:et-fi}, we show BLEU scores for 
other experiments presented in the paper only in terms of performance relative to full FT. Additionally we show adapter-1024 and X-attention scores for M2M-100; in general adapter-1024 outperforms X-attention, and both methods come close to full FT performance or slightly outperform it. Note that for M2M, for the two smallest dataset sizes (cs and vi) we see adapter-1024 (and adapter-2 for the medium size M2M) outperforming full fine-tuning, similarly to \S~\ref{sec:data-size}.

In Table~\ref{tab:all-mBART} we show results of a smaller (40m parameters) transformer model trained from scratch on each dataset separately, with an architecture consisting of 6 encoder and decoder layers, hidden dimension of 512 and feed-forward hidden dimension 1024. We train a unique sentencepiece \cite{kudo-richardson-2018-sentencepiece} vocabulary for each dataset, shared between source and target language, of size approximately 16k. Training hyper-parameters were the same as our other models. For the  \textit{x$\rightarrow$en} direction almost all of our methods based on pre-trained models outperformed the `from scratch' baseline, however in the  \textit{en$\rightarrow$x} direction for mBART the most parameter efficient methods sometimes fall short (see e.g.\ Turkish or French). For translating into Farsi no pre-trained model outperformed the from scratch model, even with full fine-tuning, suggesting a weakness for particularly low resource resource languages like Farsi. Note per-dataset hyper-parameter search would likely improve performance, especially for `from scratch' results, but we did not attempt this due to computational constraints.

\section{Additional Correlation Results for Language Distance}
\label{sec:extracorrel}

Table~\ref{tab:correlation} shows additional correlation coefficient between PEFTs' performances and different language distances: syntax, phonology and geography. Moreover, Table \ref{tab:billion-correl} shows the correlation coefficients between language distance and relative performance for the 1.2 billion size M2M model.

\begin{table*}
\small
\begin{tabular}{@{}llcccccccccc@{}}
\toprule
& No. & fa & it & de & ru & ko & fr & pt & tr & vi & cs \\
\textbf{M2M-100} & Params & 150k & 230k & 208k & 208k & 205k & 192k & 184k & 182k & 133k & 103k \\
\midrule
\noalign{\smallskip}
\textit{\textbf{en}$\rightarrow$\textbf{x}} \\
\noalign{\smallskip}
Full FT & 484m & 17.6 & 32.8 & 32.0 & 22.0 & 9.6 & 41.3 & 42.1 & 17.9 & 33.9 & 24.5 \\
Full FT (1.2b) & 1.2b & 17 & 33.7 & 33.6 & 23.8 & 9.9 & 42.8 & 43.1 & 18.8 & 35.2 & 26.3 \\
Adapter (b=1024) & 50m & 17.6 & 32.7 & 31.3 & 22.0 & 9.3 & 41.0 & 42.4 & 17.9 & 33.4 & 24.8 \\
X-attention & 50m & 17.3 & 31.7 & 31.0 & 21.9 & 9.2 & 40.8 & 42.0 & 17.7 & 33.6 & 24.5 \\
BitFit & 335k & 16.0 & 28.9 & 27.3 & 20.0 & 7.8 & 37.7 & 38.5 & 15.2 & 31.5 & 22.4 \\
Adapter (b=5) & 320k & 16.2 & 29.3 & 27.5 & 20.5 & 8.1 & 37.8 & 39.0 & 15.6 & 31.4 & 22.8 \\
Adapter (b=2; 1.2B) & 344k & 14.6 & 32.5 & 32.1 & 23.1 & 8.9 & 42.2 & 43.1 & 16.7 & 34.6 & 26.4 \\
No FT (1.2B) & 0 & 9.7 & 29.6 & 29.9 & 21.1 & 5.5 & 37.6 & 39.6 & 13.2 & 32.9 & 24.0 \\
No FT (484M) & 0 & 10.6 & 26.8 & 25.9 & 18.4 & 5.0 & 33.6 & 35.8 & 12.4 & 30.0 & 20.6 \\
\noalign{\smallskip} 
\hdashline 
\noalign{\smallskip}
\textit{\textbf{x}$\rightarrow$\textbf{en}} \\
\noalign{\smallskip}
Full FT & 484m  & 32.3 & 36.6 & 37.2 & 27.8 & 22.2 & 43.2 & 47.9 & 30.1 & 34.3 & 32.8 \\
Full FT (1.2b) & 1.2b & 36.1 & 39 & 39.3 & 29.9 & 23.9 & 45.1 & 49.8 & 31.6 & 35.7 & 35.3 \\

Adapter (b=1024) & 50m  & 32.3 & 36.3 & 36.3 & 28.0 & 22 & 43.2 & 47.8 & 30 & 34.7 & 33.9 \\
X-attention & 50m & 31.6 & 36.1 & 36.3 & 27.1 & 21.4 & 42.8 & 47.1 & 29.2 & 33.6 & 33.4 \\
BitFit & 335k  & 27.2 & 32.6 & 32.9 & 24.5 & 19.0 & 39.4 & 44 & 26.4 & 31.5 & 31.3 \\
Adapter (b=5) & 320k & 28.8 & 33.1 & 33.2 & 25.5 & 19.6 & 40.2 & 44.8 & 26.9 & 33.3 & 31.9 \\
Adapter (b=2; 1.2B) & 344k & 31.5 & 37.3 & 37.7 & 28.9 & 22.2 & 44.0 & 48.7 & 29.9 & 37.5 & 35.6 \\
No FT (1.2B) & 0 & 14.9 & 32.5 & 32.1 & 24.1 & 17.6 & 37.5 & 42.0 & 24.2 & 29.9 & 30.1 \\
No FT (484m) & 0 & 14.9 & 29.7 & 29.5 & 21.4 & 15.8 & 34.9 & 38.6 & 22.0 & 27.1 & 27.2 \\
\bottomrule
\end{tabular}
\caption{x$\leftrightarrow$en results in terms of BLEU for M2M-100 experiments.}
\label{tab:all-m2m}
\end{table*}

\begin{table*}
\small
\begin{tabular}{@{}llcccccccccc@{}}
\toprule
& No. & fa & it & de & ru & ko & fr & pt & tr & vi & cs \\
\textbf{mBART} & Params & 150k & 230k & 208k & 208k & 205k & 192k & 184k & 182k & 133k & 103k \\
\midrule
\noalign{\smallskip}
\textit{\textbf{en}$\rightarrow$\textbf{x}} \\
\noalign{\smallskip}
Full FT & 610m & 17.8 & 32.9 & 33.1 & 23.5 & 10.1 & 42.7 & 43.5 & 18.7 & 35.2 & 25.2 \\
Adapter (b=1024) & 50m & 18.0 & 33.3 & 32.8 & 22.9 & 9.9 & 37.9 & 42.8 & 18.2 & 34.6 & 24.3 \\
X-attention & 50m & 16.8 & 27.7 & 30.3 & 21.2 & 8.8 & 39.5 & 40.8 & 16.3 & 33.5 & 22.2 \\
BitFit & 335k  & 12.8 & 22.7 & 23.3 & 16.6 & 5.3 & 30.9 & 30.9 & 9.5 & 26.8 & 15.6 \\
Adapter (b=5) & 320k & 13.7 & 22.7 & 23.9 & 15.4 & 5.8 & 29.3 & 32.3 & 9.9 & 27.3 & 15.4 \\
From Scratch & 40m & 25.0 & 23.9 & 22.9 & 15.3 & 5.5 & 32.5 & 35.4 & 11.0 & 26.2 & 17.0 \\
\noalign{\smallskip} 
\hdashline 
\noalign{\smallskip}
\textit{\textbf{x}$\rightarrow$\textbf{en}} \\
\noalign{\smallskip}
Full FT & 610m & 33.9 & 38.2 & 34.1 & 29.6 & 23.5 & 44.8 & 49.4 & 31.7 & 36.0 & 34.3 \\
Adapter (b=1024) & 50m & 32.8 & 38.0 & 33.5 & 28.9 & 22.9 & 44.4 & 48.6 & 30.6 & 35.2 & 32.9 \\
X-attention & 50m& 30.8 & 32.9 & 31.6 & 26.8 & 19.7 & 41.9 & 43.8 & 27.0 & 34.0 & 31.1 \\
BitFit & 335k & 23.6 & 29.5 & 25.9 & 22.0 & 14.9 & 35.2 & 38.8 & 21.1 & 28.1 & 26.0 \\
Adapter (b=5) & 320k & 23.2 & 29.9 & 25.7 & 21.8 & 15.4 & 34.8 & 38.2 & 21.0 & 27.4 & 26.4 \\
From Scratch & 40m & 20.9 & 27.3 & 26.3 & 19.2 & 11.6 & 34.3 & 39.4 & 19.1 & 21.9 & 23.8 \\
\bottomrule
\end{tabular}
\caption{x$\leftrightarrow$en results in terms BLEU for mBART experiments.}
\label{tab:all-mBART}
\end{table*}

\begin{table*}[]
\centering
\small
\begin{tabular}{@{}lcccccc|cccccc@{}}
\toprule
& \multicolumn{6}{c}{\textbf{M2M-100}} & \multicolumn{6}{c}{\textbf{mBART}} \\
& en < & > fi & en < & > et &  fi < & > et & en < & > fi & en < & > et & fi < & > et \\ \midrule
Full FT             & 43.9            & 37.9  & 40.4            & 33.4  & 33.6  & 33.4  & 45.4            & 39.8            & 42.3            & 35.5            & 34.8           & 35.4            \\
Adapter (b=1024)   & 42.7            & 35.5  & 39.6            & 30.9  & 31.6  & 31.5  & 45.3            & 39.1            & 41.9            & 33.8            & 33.6           & 33.8            \\
X-attention     & 42.9            & 35.9  & 39.5            & 31.2  & 31.6  & 31.1  & 40.6            & 34.2            & 36.1            & 28.9            & 28.5           & 29.1            \\
BitFit      &  35.4            & 25.6  & 33.9            & 22.9  & 26.8  & 26.1  & 28.9            & 18.9            & 25.0                & 13.8            & 18.0            & 17.3            \\
Adapter (b=5)    & 36.1            & 26.8  & 34.3            & 23.2  & 26.9  & 26.5  & 28.9            & 19.2            & 24.3            & 14.6            & 18.2           & 16.9            \\
Adapter (b=2; 1.2B)        & 41.9            & 32.0      & 39.6            & 28.8  & 31.8  & 31.4  & - & - & - & - &- &    -             \\
No FT (1.2B)                & 40.3            & 28.6  & 38.1            & 27.3  & 31.3  & 31.0    & - & - & - & - &- &     -            \\
No FT (484M)               & 34.1            & 23.6  & 32.9            & 22.6  & 26.8  & 26.2  & - & - & - & - &- &    -             \\ \bottomrule
\end{tabular}
\caption{(en, et, fi) results in terms of BLEU for M2M-100 and mBART experiments. Note that BLEU scores are not directly comparable as the datasets are different for each language-pair. For a comparison between fine-tuning methods, we refer to relative performances over full fine-tuning (Fig.~\ref{fig:into-fi/et}).}
\label{tab:et-fi}
\end{table*}

\end{document}